\documentclass[preprints,article,accept,pdftex,moreauthors]{Definitions/mdpi} 
\usepackage[final]{changes}

\firstpage{1} 
\pubvolume{1}
\issuenum{1}
\articlenumber{0}
\pubyear{2023}
\copyrightyear{2023}
\datereceived{} 
\dateaccepted{} 
\datepublished{} 
\hreflink{https://doi.org/} % If needed use \linebreak
\usepackage{subcaption}
\usepackage{comment}
\usepackage[nolist,nohyperlinks]{acronym}

\acrodef{rdspp}[RDS-PP\textsuperscript{\textregistered}]{Reference Designation System for Power Plants\textsuperscript{\textregistered}}
\acrodef{rdsps}[RDS-PS]{Reference Designation System for Power Systems}
\acrodef{zeus}[ZEUS]{State-Event-Cause-Key (in German "Zustand-Eregnis-Ursachen-Schlüssel")}
\acrodef{digma}[digitalization workflow]{workflow for digitalization of maintenance information}
\acrodef{ocr}[OCR]{Optical Character Recognition}
\acrodef{wea}[WEA]{Windenergieanlage}
\acrodef{csv}[CSV]{comma seperated value}
\acrodef{nlp}[NLP]{Natural Language Processing}
\acrodef{kpi}[KPI]{Key Performance Indicator}
\acrodef{tlp}[TLP]{Technical Language Processing}
\acrodef{svm}[SVM]{Support Vector Machine}
\acrodef{lcoe}[LCOE]{\emph{Levelized Cost of Energy}}
\acrodef{om}[O\&M]{\emph{Operation and Maintenance}}
\acrodef{wt}[WT]{\emph{Wind Turbine}}
\acrodef{wf}[WF]{\emph{Wind Farm}}
\acrodef{ai}[AI]{\emph{Artificial Intelligence}}
\acrodef{se}[SE]{\emph{Service Enterprise}}
\acrodef{tdomm}[TDoMM]{text description of the maintenance measure}
\acrodef{erp}[ERP]{enterprise resource planning}
\acrodef{mwo}[MWO]{Maintenance Work Order}
\acrodef{wmep}[WMEP]{Scientific measurement and evaluation program (in German "Wissenschaftliches Mess- und Evaluierungsprogramm")}
\acrodef{mtbf}[MTBF]{Mean Time Between Failure}
\acrodef{ml}[ML]{Machine Learning}
\acrodef{nb}[NB]{Naive Bayes}
\acrodef{lr}[LR]{Logistic Regression}
\acrodef{smote}[SMOTE]{Synthetic Minority Oversampling Technique}
\acrodef{ro}[RO]{Random Oversampling}

\Title{KPI Extraction from Maintenance Work Orders --- A comparison of Expert Labeling, Text Classification and AI-Assisted Tagging for Computing Failure Rates of Wind Turbines}

\TitleCitation{Title}

\Author{Marc-Alexander Lutz $^{1}$\orcidA{}, Bastian Schäfermeier $^{1}$* \orcidD{}, Rachael Sexton$^{2}$ \orcidC{}, Michael Sharp $^{2}$ \orcidE{}, Alden Dima $^{2}$  \orcidB{}, Stefan Faulstich $^{1}$ \orcidF{} and Jagan~Mohini Aluri $^{1}$}

\AuthorNames{Firstname Lastname, Firstname Lastname and Firstname Lastname}

\AuthorCitation{Lastname, F.; Lastname, F.; Lastname, F.}
\address{%
$^{1}$ \quad Fraunhofer Institute for Energy Economics and Energy System Technology, Joseph-Beuys-Straße 8,
34117 Kassel, Germany\\
$^{2}$ \quad National Institute for Standards and Technology, 100 Bureau Drive, Gaithersburg, MD 20899, United States of America}

\corres{Correspondence: bastian.schaefermeier@iee.fraunhofer.de}

\abstract{Maintenance work orders are commonly used to document information about wind turbine operation and maintenance. This includes details about proactive and reactive wind turbine downtimes, such as preventative and corrective maintenance. However, the information contained in maintenance work orders is often unstructured and difficult to analyze, \replaced{presenting challenges}{making it challenging} for decision-makers \added{wishing} to use \replaced{it}{this information} for optimizing operation and maintenance. To address this issue, this work \deleted{presents}\added{compares} three different approaches to calculate reliability key performance indicators from maintenance work orders. The first approach involves manual labeling of the maintenance work orders by domain experts, using the schema defined in an industrial guideline to assign the label accordingly. The second approach involves the development of a model that automatically labels the maintenance work orders using text classification methods. \added{Through this method, we are able to achieve macro average and weighted average $F_1$-Scores of 0.75 and 0.85 respectively,} The third technique uses an AI-assisted tagging tool to tag and structure the raw maintenance information\deleted{contained in the}\added{, together with a novel rule-based approach for extracting relevant maintenance work orders for failure rate calculation} . \deleted{The resulting calculated reliability key performance indicator of the first approach \replaced{is}{are} used as a benchmark for comparison with the results of the second and third approaches. The quality and time spent are considered as criteria for evaluation.} \added{In our experiments the AI-assisted tool leads to a 88\% drop in tagging time in comparison to the other two approaches, while expert labeling and text classification are more accurate in KPI extraction. Overall, \deleted{these three methods} \added{our findings} make extracting  maintenance information from maintenance work orders more efficient, enable the assessment of reliability key performance indicators and therefore support the optimization of wind turbine operation and maintenance.}}


\begin{document}

%%%%%%%%%%%%%%%%%%%%%%%%%%%%%%%%%%%%%%%%%%
\setcounter{section}{0} %% Remove this when starting to work on the template.

\section{Introduction}
\ac{om} represent around 25 percent of the overall cost in the life cycle of a \ac{wt}. After commissioning costs related to maintenance are often the only variable cost \cite{InternationalRenewableEnergyAgency.}. Ways to reduce those variable costs therefore need to be addressed. Costs of \ac{om} can be divided into two main categories: Proactive maintenance and reactive maintenance. Proactive maintenance, also known as preventive maintenance, is scheduled in advance and is intended to prevent equipment failure. Reactive maintenance, also known as corrective maintenance, is conversely performed in response to unpredicted or observed deterioration of a system, such as when equipment fails and needs to be repaired \deleted{in order}to restore it to normal working condition.

One of the challenges related to \ac{om} costs of \acp{wt} is the lack of standardization in the acquisition,  communication\added{,} and the documentation of maintenance information, e.g. in the form of \acp{mwo} or service reports. \acp{mwo} are often prepared by different service technicians\deleted{,} and can vary in both the level of detail provided and the format used. This makes it difficult to perform reliability analysis and subsequently to effectively plan and schedule maintenance activities. In order to reduce \ac{om} costs, it is important to develop standardized methods for acquisition, communication and documentation of  \acp{mwo} and to use the information in the large quantity of \acp{mwo}  for better understanding of  faults and failures and for optimization of \ac{om} itself.

\ac{wt} \acp{mwo} are often unstructured and difficult to use for reliability analysis and decision-making \cite{lutz2022digitalization}. This research explores three separate methods for investigating and extracting reliability \ac{kpi} from the large volume of \acp{mwo} created in the WT industry. Our goal is to enable efficient and effective use of this underutilized source of information through both human-in-the-loop procedures and computer accelerated tools.

The first approach involves manual labeling of textual \acp{mwo} according to the technical guideline \ac{zeus} \cite{FGWe.V.FordergesellschaftWindenergieundandereErneuerbareEnergien.2012}. In the second approach,  models are developed to automate the classification of \acp{mwo} according to the \ac{zeus} guideline. The third approach involves the usage of a tool to tag \acp{mwo}.
Once the \acp{mwo} have been classified using these methods,  reliability \acp{kpi} can be calculated for each. The results are then compared and assessed qualitatively and quantitatively to determine which approach is the most effective at generating reliable and useful \ac{kpi}.

\added{The research questions of this work can be summarized as follows:
%\begin{itemize}
%\item 
1. How can manual classification, machine-learning-based text classification and AI-Assisted tagging be employed for KPI-prediction from maintenance work orders?
%performance and manual effort?
%\item 
2. How do the approaches compare against each other in terms of KPI prediction performance and manual effort and what are the individual strengths and weaknesses?}
%3. How can be dealt with imbalanced classes in maintenance work order classification?}
%\end{itemize}

\added{
The main contributions of this work are as follows: 1. We give a comparison of KPI-extraction from wind turbine maintenance work orders based on first, the AI-assisted tagging tool ``Nestor'' \cite{Brundage.2021} and second, automated text classification through logistic regression and naive Bayes. To the best of our knowledge, the present work is the first to give a hands-on comparison of text classification and AI-based tagging for \ac{mwo} KPI extraction. 2. For the tagging method, we introduce a novel tagging approach for tagging and extracting \acp{mwo} relevant for failure rate calculation. Through this process, we are able to save almost 90\% of the effort when compared to completely manual document labeling. 3. Based on an evaluation of the different methods concerning classification performance and manual effort, we document strengths and weaknesses of the individual methods and give recommendations for their future application. We identify common pitfalls with maintenance work order classification, namely class imbalance and lack of training data. Our results show that different oversampling techniques are able to mitigate such problems to a different degree.}

This research paper is structured as follows:

\begin{itemize}
    \item Section \ref{sota}: This section explores state-of-the-art research on existing initiatives that assess  reliability \ac{kpi} in \ac{wt} \ac{om}. This includes a review of tools for tagging e.g. textual maintenance data. Lastly, \deleted{a }research on existing \acp{kpi} in \ac{wt} and other industries.
    \item Section \ref{methods}: This section outlines the methodology of the three different approaches for structuring the \acp{mwo} of \acp{wt} to extract reliability \acp{kpi}. The \added{\ac{kpi} }calculation \deleted{of the \acp{kpi} }is outlined here. We demonstrate how they compare against each other.
    \item Section \ref{results}: The methods described in the  Section \ref{methods} are applied to the \acp{mwo} of an offshore \ac{wf}. The resulting \ac{kpi} values are presented and compared.
    \item Section \ref{discussion}: The results are discussed and compared to the results from other initiatives presented in section \ref{sota}.
    \item Section \ref{conclusion}: This section concludes this research work and highlights future areas of research based on this work.
    \end{itemize}

\section{State of Research in Knowledge Discovery in Maintenance Work Orders}
\label{sota}
In this section an overview of the state of research is given.
First, relevant initiatives are presented which collected reliability \ac{kpi} of \acp{wt}. However, those \acp{kpi} are not extracted \replaced{from}{out of} \ac{mwo} but are stated as reference for the discussion of the results in section \ref{discussion}.
Second, research in  knowledge discovery  from \acp{mwo} is outlined. \replaced{Other}{Different} methods and \deleted{other} approaches from different domain\added{s} are listed.
Third, existing results in  literature are shown with regards to \acp{kpi} based on \acp{mwo} of several types of machinery and \acp{wt}. 

\subsection{Relevant Initiatives and their Reliability \ac{kpi} for Wind Turbines}
Much research \deleted{work }is available \replaced{for}{in the} \added{\ac{wt}} reliability assessment\deleted{ of \acp{wt}}. 
\replaced{\citet{Pfaffel.2017} compared}{A comparison of} different initiatives \replaced{and}{which }collected \ac{om} information to assess reliability \acp{kpi}\deleted{is made in \citet{Pfaffel.2017} }. A review of data related to reliability, availability\added{,} and maintenance for the identification of trends in offshore \acp{wt} can be seen in  \citet{CEVASCO2021110414}.
Two other sources are briefly highlighted in this subsection, namely the \ac{wmep} \cite{echavarria2008reliability} and  \citet{carroll2016failure}, where \acp{kpi} such as the failure rate are publicized.

\subsubsection{WMEP}
The \ac{wmep}-program carried out by Fraunhofer IWES investigated \added{reported} maintenance data \replaced{reported by about}{from} 1500 \acp{wt}. \replaced{It included}{The maintenance data consists of} \added{failure and operational performance} details \deleted{about failures, operational performance and more}\replaced{ and served as }{. It is} a comprehensive collection of reliability data, with more than sixty thousand reports on maintenance and repair measures. These reports also \replaced{contain}{consist of} technical details and information about \ac{om} costs. For the\added{se} \acp{wt} \deleted{considered}, the program yielded a failure rate of 2.60~1/a \cite{echavarria2008reliability}. 

\subsubsection{Strathclyde}
The University of Strathclyde \deleted{performed}research\added{ed}\deleted{ on} data collected from\deleted{ about 350} \acp{wt}\deleted{,}\replaced{and provided}{providing} failure rates and other \acp{kpi}.  They analyzed and categorized each failure as \replaced{a}{either} major repair, \added{a} minor repair or \added{a} major replacement. For the\replaced{se }{considered} \acp{wt},  \citet{carroll2016failure} \replaced{reported}{yielded} a failure rate of around 8.27~1/a.

\subsection{Knowledge Discovery using MWOs}
Different procedures and methods exist to extract knowledge from textual descriptions of \acp{mwo}. The most basic\deleted{way to} extract the knowledge \deleted{is} through manual labeling and analysis. Text mining techniques like classification and clustering can also be employed to extract failure data from \deleted{the }work orders \cite{arif2017extracting}. \deleted{A work from }Kazi et al. \cite{arif2017extracting} propose\deleted{s} a methodology for extracting failure data from \deleted{the }\acp{mwo} based on \deleted{its }downtime data and develop\deleted{s} a classifier that \replaced{associates}{classifies} a downtime event \replaced{with}{into} one of two classes: \replaced{n}{N}onfailure or failure. \ac{ml} models are trained through supervised learning with large datasets. Processing and learning from the raw data directly does not guarantee \added{that} the correct patterns or trends will be captured or learned by the models, so training data must be labeled. Training data is tagged with target attributes that relate to the overall goal of the model as well as the intrinsic nature of the data entries to allow the \ac{ml} agent to \deleted{more} quickly and accurately converge on desirable trends useful for decision making \cite{monarch2021human}. This tagging process is performed \added{either} manually or through the use of tools. Using  Technical Language Processing techniques\cite{Brundage.2021}, relevant information can be extracted, such as the specific equipment or component mentioned in the \acp{mwo}, the nature of the maintenance task being performed, and any notes or instructions related to the work. Automating the maintenance process, tracking maintenance history, and performing predictive maintenance can then be achieved with this information \cite{Brundage.2021}.

\subsection{\ac{kpi} using Knowledge Discovery}
Monitoring and extracting \acp{kpi} is important for enterprises and industries to provide insight into their maintenance processes. Intuitively, \acp{mwo} contain much \added{\ac{kpi}-related} information, either numeric or text based, \deleted{relating to these \acp{kpi} }that can be extracted through the proper tools and procedures. Some of the \acp{mwo}-based reliability \acp{kpi} are Machine per Time Between Failure, Machine by Problem Action per Time Between Failure, etc. \cite{brundage2018developing}. Traditionally, the analysis of this information \replaced{was}{had to be} done manually, requiring many man-hours and intimate \deleted{levels of} expert knowledge of both the system and the domain. More recently this task has been accelerated through the use of computer aided tools and AI technologies that can automatically extract, process, and relate this information to user selected \acp{kpi}. These more recent tools and techniques provide unique opportunities in the realms of maintenance and reliability, but are far from perfect. They face a multitude of challenges given the imperfect nature of real world data collection. Text entries in the \acp{mwo} are often prone to misspellings, unique abbreviations, shorthand, etc \cite{Navinchandran.2021}. \deleted{A work from }Mukherjee and Chakraborty present\deleted{s} an unstructured text analysis method by developing diagnostic fault trees from historic maintenance logs of a power plant \cite{mukherjee2007automated}. \deleted{Another work from }Navinchandran \cite{Navinchandran.2021} \added{also} present\deleted{s} a systematic workflow for discovering the right \acp{kpi} from the \acp{mwo} that help perform \deleted{a} sensitivity-type \replaced{analyses}{analysis} \deleted{in order} to determine the significance and influence of \deleted{the} identified concepts, which can then be interpreted within the facility context. 
% add chunk on measuring impact of data quality on kpi estimation? could also leave for section 2.1.3, or a discussion topic if e.g. data quality is out of scope for now.
Lutz et al. present a digitalization workflow in the wind energy domain, to extract and structure maintenance data as a basis for  reliability KPI calculation \cite{lutz2022digitalization}. \deleted{A work from }Frank et al. \cite{torres2020indicator} present\deleted{s} a methodology that uses \acp{kpi} and key risk indicators to assess the safety and security of offshore wind farms.

\section{Methods}
\label{methods}

\begin{figure}
    \centering
    \includegraphics[width=\linewidth]{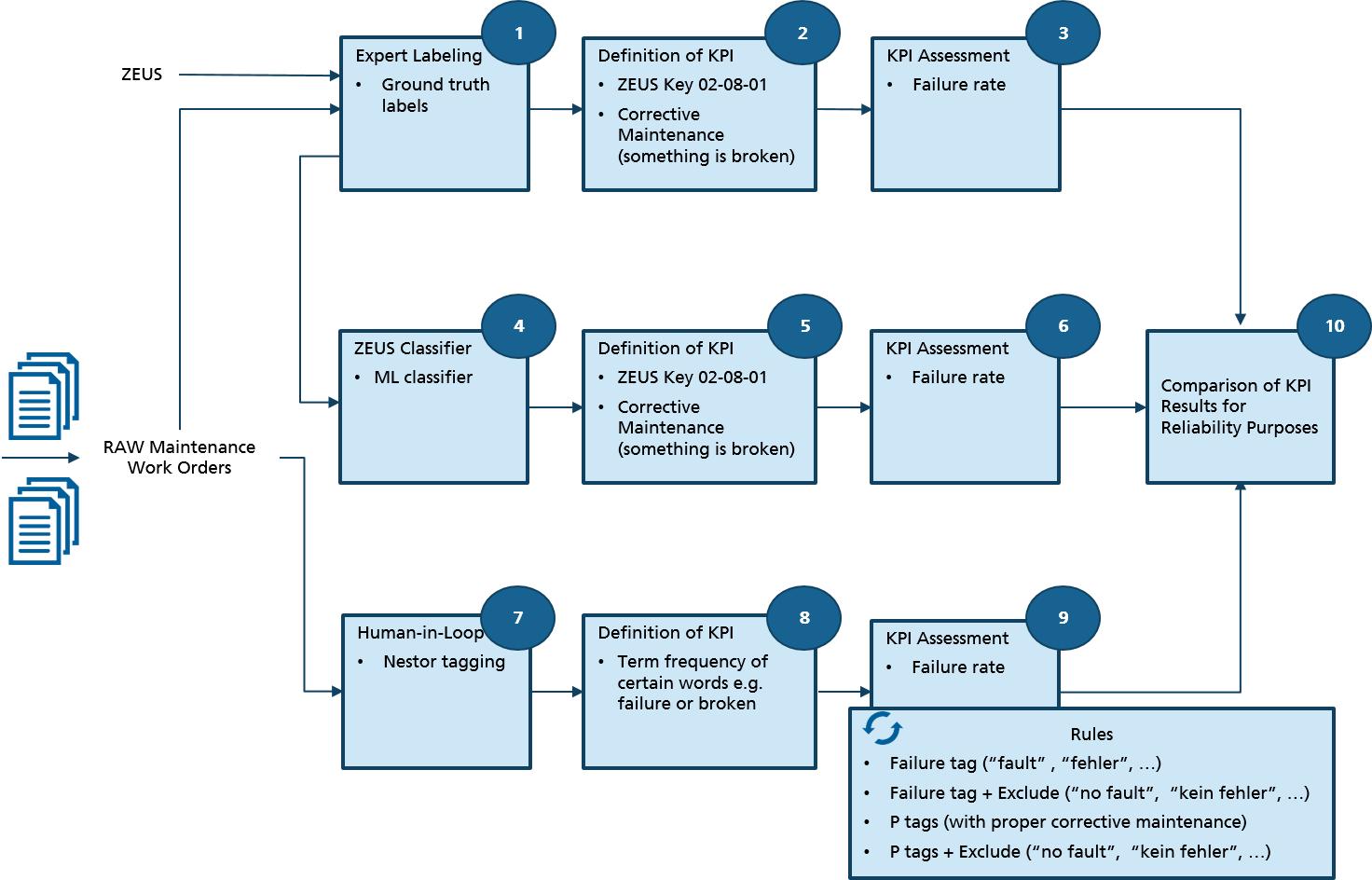}
    \caption{\added{Visualization of our} methodology. \added{We compare three different approaches: 1. Expert labeling (first row). 2. Machine-Learning-based classification (second row) 3. Human-in-the-loop tagging (third row).}}
    \label{fig:Methodology}
\end{figure}

\replaced{We implement t}{T}hree different approaches \deleted{are implemented} to extract reliability \acp{kpi} \replaced{from}{out of} \acp{mwo}. \replaced{We then compare}{Thereafter,} the resulting \acp{kpi} \deleted{are compared with each other}. The approaches and their comparison are visualized in Figure \ref{fig:Methodology}. The first approach is \deleted{referred to as} expert labeling. It is seen in the first row indicated by the boxes 1, 2 and 3. The steps of the expert labeling approach are as follows: The raw \acp{mwo} are manually labeled according to the \ac{zeus} technical guideline by domain experts (box 1 in Figure \ref{fig:Methodology}). The technical guideline is more elaborately described in section \ref{subsubsec:zeus1}. Subsequently relevant \acp{kpi} are defined and the defined \acp{kpi} are assessed (box 2 and box 3, respectively in Figure \ref{fig:Methodology})\\

The second approach\added{, ML-based text classification,} is shown in the second row and is indicated by the boxes 4, 5 and 6. \deleted{It is referred to as automated classification.} Instead of manually labeling  \acp{mwo} according to a technical guideline a classifier is trained \replaced{which}{to} automatically predict\deleted{s} the corresponding label. This is indicated in box 4 in Figure \ref{fig:Methodology}. \replaced{Like}{Similarly as in} the expert labeling method\added{,} \acp{kpi} are defined and assessed (box 5 and box 6\added{,} respectively\added{,} in Figure \ref{fig:Methodology})\\

The boxes 7, 8 and 9 are related to the third approach\replaced{, }{. It is referred to as} human in the loop tagging. The tool used to tag the raw \acp{mwo} is described in detail in \ref{subsubsec:tagging tool}. The \ac{kpi} definition and assessment for this approach are indicated by box 8 and 9 respectively. Box 10 indicates a comparison of \ac{kpi} results from the three different methods.

\subsection{Expert Labeling}
Manual labeling refers to the process of labeling \acp{mwo} by the domain experts familiar with usage of the \ac{zeus} technical guideline. Each \ac{mwo} is manually assigned to a  label according to the \ac{zeus} technical guideline by the experts, based on the information contained in the text descriptions in the \acp{mwo}. These tagged samples are referred to as ground truth labels.
\replaced{We further describe}{To further outline the process of }the expert labeling approach in the next section\added{, where} the \ac{zeus} technical guideline is outlined. 

\subsubsection{Technical Guideline: \ac{zeus}}
\label{subsubsec:zeus1}

Each state of the \ac{wt} can be defined in a uniform \replaced{fashion}{matter} by using \ac{zeus} \cite{FGWe.V.FordergesellschaftWindenergieundandereErneuerbareEnergien.2012}. Within \ac{zeus}  several blocks are defined. Each block \replaced{asks}{raises} a question about the state of the turbine or \replaced{component}{the considered element}. By answering the question an identifier is given. The set of all identifiers describe the state of the \ac{wt} and the considered element in a uniform way.
In this paper only the \ac{zeus} block 02-08 is used. It raises the question: "Which maintenance type is active or will be necessary to eliminate a deviation from the target condition?" The answers and the identifiers, namely the \ac{zeus} code, can be seen in Table \ref{tab:zeus block 02-08}

% Table generated by Excel2LaTeX from sheet 'Tabelle2'
\begin{table}
  \centering
  \caption{\added{Overview of the \ac{zeus} classification system \cite{FGWe.V.FordergesellschaftWindenergieundandereErneuerbareEnergien.2012}. Codes, levels and names are given for \ac{zeus} block 02-08, which was considered for \ac{mwo} classification in this work.}}
    \begin{tabular}{lrl}
    \toprule
    \multicolumn{1}{l}{\textbf{ZEUS CODE}} & \multicolumn{1}{l}{\textbf{ZEUS LEVEL}} & \textbf{ZEUS NAME} \\
    \midrule
    \textbf{02-08} & \textbf{2} & \textbf{maintenance type} \\
    \midrule
    02-08-01 & 3     & corrective maintenance \\
    02-08-01-01 & 4     & deferred corrective maintenance \\
    02-08-01-02 & 4     & immediate corrective maintenance \\
    \midrule
    02-08-02 & 3     & preventive maintenance \\
    02-08-02-01 & 4     & predetermined maintenance \\
    02-08-02-02 & 4     & condition based maintenance \\
    02-08-02-03 & 4     & predictive maintenance \\
    \midrule
    02-08-97 & 3     & undefined maintenance type \\
    02-08-96 & 3     & unresolved maintenance type \\
    02-08-XX & 3     & insignificant attribute \\
    \bottomrule
    \end{tabular}%
  \label{tab:zeus block 02-08}%
\end{table}%

\subsubsection{Definition of \acp{kpi}}
\label{subsubsec:definitionofkpi1}
Although there are five classes in the third level of \ac{zeus} block 02-08, only the \acp{mwo} that belong to the class 02-08-01 are considered for failure rate \ac{kpi} calculation as this class indicates corrective maintenance activities that aim at fixing defects in a \ac{wt} equipment in contrast to, e.g., preventive or predictive maintenance measures.

\subsubsection{\acp{kpi} assessment}
\label{subsubsec:kpiassessment1}
The failure rate calculation involves a series of steps. \added{As the first step} \ac{mtbf} is calculated for \replaced{each}{the} \ac{wt} (cf. equation \ref{eu_mtbf}). \added{We assume that the maintenance event rate for the corrective maintenance \ac{zeus} class can be considered as the failure rate of the \ac{wt}. Hence, we only consider corrective maintenance events as failures for \ac{mtbf} calculation, which in turn are determined from the \acp{mwo} corresponding to the \ac{wt}.} In the next step, the maintenance event rate is obtained by calculating the inverse of the \ac{mtbf} values as given in equation \ref{eu_failurerate}. \deleted{We assume that the maintenance event rate for the corrective maintenance class can be considered as the failure rate of the \ac{wt}.}
\added{We report the average failure rate over all \acp{wt} in our data. Intuitively, this number gives us the average number of corrective maintenances performed per year per wind facility.}

\begin{equation} \label{eu_mtbf}
MTBF_{Item} = \frac{ \sum_{i=1}^{C_{F,Item}} \Delta t_{Item,i}} {C_{F,Item}}
\end{equation}

\begin{equation} \label{eu_failurerate}
\lambda_{item} = \frac{1}{MTBF_{item}} 
\end{equation}

Where,
\[MTBF_{Item} =  \textnormal{Mean time between failure per item \added{(i.e., in our case, per \ac{wt})},}\]
\[\lambda_{item} =  \textnormal{Failure rate per item,}\]
\[\Delta t_{Item,i} = \textnormal{Time to \emph{ith} failure \deleted{in years},}\]
\[C_{F,Item} = \textnormal{Count of failures per item}\]

\begin{table}
  \centering
  \caption{\added{Example maintenance work orders. We compute failure rates for each \ac{wt} based on the corresponding maintenance work orders for ZEUS Code 02-08-02 (corrective maintenance). Note that \ac{zeus} codes are not part of the original \acp{mwo} but must be derived, e.g., through manual expert labeling or automated text classification.}}
  \label{tab:mwoexample}
    \begin{tabular}{llp{7.5cm}r}
    \toprule
\textbf{Start Date} & \textbf{WT} & \textbf{Description} & \textbf{ZEUS CODE} \\
    \midrule
    2023-08-01 & WT1 & Emergency generator refilled with diesel. & 02-08-02\\
    2023-08-10 & WT2 & Internal blade inspection.& 02-08-02\\
    2023-09-01 & WT1 & Troubleshooting at crane on outside platform performed.
Thermo relay exchanged. & 02-08-01\\
    2023-09-03 & WT2 & Hydraulic hoses exchanged. Additional service required. & 02-08-01\\
    2023-09-25 & WT1 & Pitch batteries exchanged at axle 2. & 02-08-01\\
    2023-10-27 & WT2 & Grommets are in position. No correction needed. & 02-08-97\\
    2023-10-14 & WT1 & Fixed connector cable. & 02-08-01\\
    2023-11-30 & WT1 & New reflector pipe installed. & 02-08-01\\
    \bottomrule
    \end{tabular}%
\end{table}%

\added{Table \ref{tab:mwoexample} contains a number of fictitious example \acp{mwo}. The failure rate for WT1 in this example can be computed by first, extracting the four \acp{mwo} with \ac{zeus} Code 02-08-01 (corrective maintenance). The time to the first failure $\Delta t_{WT1,1}$ is the time difference between the first corrective maintenance and the first day of operation of WT1. In this example, we assume $\Delta t_{WT1,i}= 90 days$. The other time deltas are the number of days between each subsequent corrective maintenance, which can be inferred from the \ac{mwo} dates. Hence, we obtain \[ MTBF_{WT1} = \frac{ \sum_{i=1}^{C_{F,WT1}} \Delta t_{WT1,i}} {C_{F,WT1}} = \frac{(90 + 24 + 19 + 47)d} {4} = 45d. \] The failure rate of WT1 thus is $\lambda_{WT1} = \frac{1}{45d} \approx 8.1/a$, i.e., about 8 failures per year. We may note that, in theory, through the same process as exemplified above, we may compute failure rates for other events than corrective maintenances, such as preventive maintenances, simply by considering \acp{mwo} from the corresponding \ac{zeus} class. Similarly, instead of considering a wind turbines as the \emph{item} in equation \ref{eu_mtbf} and \ref{eu_failurerate}, we may calculate the \acp{kpi} for wind turbine components, e.g., through \ac{mwo} classification according to the RDS-PP standard \cite{rdspp}.}

\subsection{Automated Classification according to technical guidelines}
In this section the second approach is outlined. \deleted{Therefore,  } \added{In this approach} \ac{ml}-based text classifiers are \deleted{built}\added{used} to automatically label and classify the \acp{mwo} into their respective \ac{zeus} classes. We use two supervised \acp{ml} techniques, \ac{nb} \cite{naivebayes} and \ac{lr} \cite{logregression}, to learn from the training data and classify the test set into their respective \ac{zeus} classes based on the knowledge gained after training. The dataset used for training the \ac{ml} model is imbalanced because the majority (> 90\%) of the total datapoints belong to just the 02-08-01 and 02-08-02 classes. \added{To mitigate this problem,} the data is oversampled using two techniques, namely the \ac{smote} \cite{smote} and \ac{ro}  \cite{he2009learning}. \added{In combination, this results} \deleted{These approaches result} in four \ac{ml} models: \ac{nb}+\ac{smote}, \ac{nb}+\ac{ro}, \ac{lr}+\ac{smote} and \ac{lr}+\ac{ro}. Ground truth data labeled by the domain experts is required for training the \ac{ml} model. However, because there is no human intervention in the training and classification, the classification of additional, unlabeled data occurs automatically, saving significant manual tagging effort. 

For the tagged data resulting from this method, the failure rate \ac{kpi} is calculated. The calculation of failure rate involves the same steps \deleted{that are} mentioned in the first approach\added{, i.e.,} \deleted{which involves} calculation of \ac{mtbf} and maintenance event rates.

\begin{comment}

\subsubsection{Technical Guideline: \ac{zeus}}
\label{subsubsec:zeus2}
The same technical guideline is used as stated in section \ref{subsubsec:zeus1}. Instead of manually assigning a label a classifier is trained to predict the corresponding label.

\subsubsection{Definition of \acp{kpi}}
\label{subsubsec:definitionofkpi2}
The \acp{kpi} are defined similarly as described in section \ref{subsubsec:definitionofkpi1}.

\subsubsection{\acp{kpi} assessment}
\label{subsubsec:kpiassessment2}
The \acp{kpi} are assessed similarly as described in section \ref{subsubsec:kpiassessment1}.
\end{comment}

\subsection{Human in the Loop Tagging}\label{subsubsec:tagging tool}
\deleted{This is the} \added{In the} third approach\deleted{. Here} an AI-assisted human-in-the-loop tagging tool called \emph{Nestor} \cite{Brundage.2021} is used for tagging the \acp{mwo}.\footnote{Documentation available at \url{https://nestor.readthedocs.io/en/latest/how_to_guide/using_nestor.html}.}
Nestor reads the raw maintenance data from a CSV file.\deleted{, extracts terms and sorts them in descending order of their Term Frequency - Inverse Document Frequency (TF-IDF) score} Through a user interface users can manually tag and group similar terms. \added{The efficiency of the tagging process is ensured by presenting only the most relevant terms, as computed through their respective Term Frequency - Inverse Document Frequency (TF-IDF) scores \cite{tfidf}. Furthermore, for each such term, similar terms are suggested automatically, which thus may be grouped together without much effort and tagged with a common ``alias'' term (e.g., ``failure'', ``fault'', and ``error'', which together may be mapped to the alias ``failure'').}

\deleted{The tagging does not follow the \ac{zeus} guideline as} \replaced{Additionally to the above alias term tagging, Nestor}{it} offers users an option to assign one of the following entities to the tagged words - ``Problem'', ``Solution'', ``Item'', ``Ambiguous'' and ``Irrelevant''. The named entity ``Problem'' (P) indicates some kind of fault or error in an equipment. The entity ``Solution'' (S) is assigned to the word if it indicates some form of solution to the problem. A word can also indicate an item and in that case, it is assigned the entity ``Item'' (I). The entity ``Irrelevant'' (X) is assigned to a word if it indicates a stop word or if it is not very significant. Finally, words that take different forms are assigned the entity ``Ambiguous'' (U) meaning their  \replaced{semantics depend on the context within the \ac{mwo} (i.e., homonyms).}{usage is unknown.}
%\added{Next to tagging words (and their similar words) in a group The overall tagging process through Nestor allows for training Named Entity recognition (NER) models, which, in turn, may be used for information extraction or KPI extraction. However, as done in the present work, the tagged information may also be used directly.}

For human-in-the-loop tagging, the calculation of failure rates does not follow the same steps as in approaches 1 and 2 as Nestor does not follow the \ac{zeus} guideline for tagging the \acp{mwo}. 
\added{Instead, based on the Nestor tags described above, we define a set of four rules that are used to extract those \acp{mwo}, which are considered for failure rate calculation (cf.} \deleted{Hence, as part of } step 9 in Figure \ref{fig:Methodology}):\deleted{, four different rules are considered for the calculation of the failure rate \ac{kpi}.} In the first rule, we ensure that only \acp{mwo} containing a ``failure'' tag are considered\deleted{ for \ac{kpi} calculation}. There are chances that the \acp{mwo} text description contains the failure tag, but the usage of this term indicates absence of a fault or a failure \replaced{due to a negation, e.g.,}{. E.g. the text description states} ''no failure detected''. As Nestor only considers the tagged word instead of the meaning of the text description, the second rule \replaced{excludes such \acp{mwo}.}{is introduced to get rid of such rows where there is no fault.} \deleted{Therefore text descriptions that state e.g. ''no failure detected'' are removed from \ac{kpi} calculation. This results in a lower failure rate value for rule 2.} Rule 1 and 2 only consider the topmost tag i.e, the word ''fault'' for \ac{kpi} calculation. However, there are \replaced{further}{also other} tags in the ''Problem'' entity that can be related to corrective maintenance\replaced{, e.g.,}{.  E.g.} ''defective component replaced'' or ''replacement of broken sensor''. Thus the third rule is introduced to consider all \deleted{the} relevant tags belonging to the P entity  for failure rate calculation. Finally, in rule 4\added{, similar to rule 2,} descriptions which contain P entity tags but which indicate absence of an error are eliminated from the failure rate calculation, e.g., "no broken sensor found. \ac{wt} back in operation".

\begin{comment}

\subsubsection{Definition of \acp{kpi}}
\label{subsubsec:definitionofkpi3}
The \acp{kpi} are defined similarly as described in section \ref{subsubsec:definitionofkpi1}.

\subsubsection{\acp{kpi} assessment}
\label{subsubsec:kpiassessment3}
The \acp{kpi} are assessed similarly as described in section \ref{subsubsec:kpiassessment1}.
\end{comment}

\subsection{Comparison of Reliability \acp{kpi}}
Finally in the last step (step 10 in Figure \ref{fig:Methodology}), the failure rate \ac{kpi} results from all three approaches are compared. Because expert labeling is manually carried out by wind industry domain experts, the \ac{kpi} resulting from the first approach are referred to as ground truth labels and they are used for benchmarking purposes. The failure rate values resulting from the second and the third techniques are compared against the \ac{kpi} results of the first method in order to assess the tagging quality. Furthermore, in each of the three approaches, the total time taken for tagging is recorded to compare the speed of tagging the \acp{mwo} data.

\section{Data Set Description}
\label{sec: Data Set Used}
\begin{figure}
    \centering
    \includegraphics[width=0.7\linewidth]{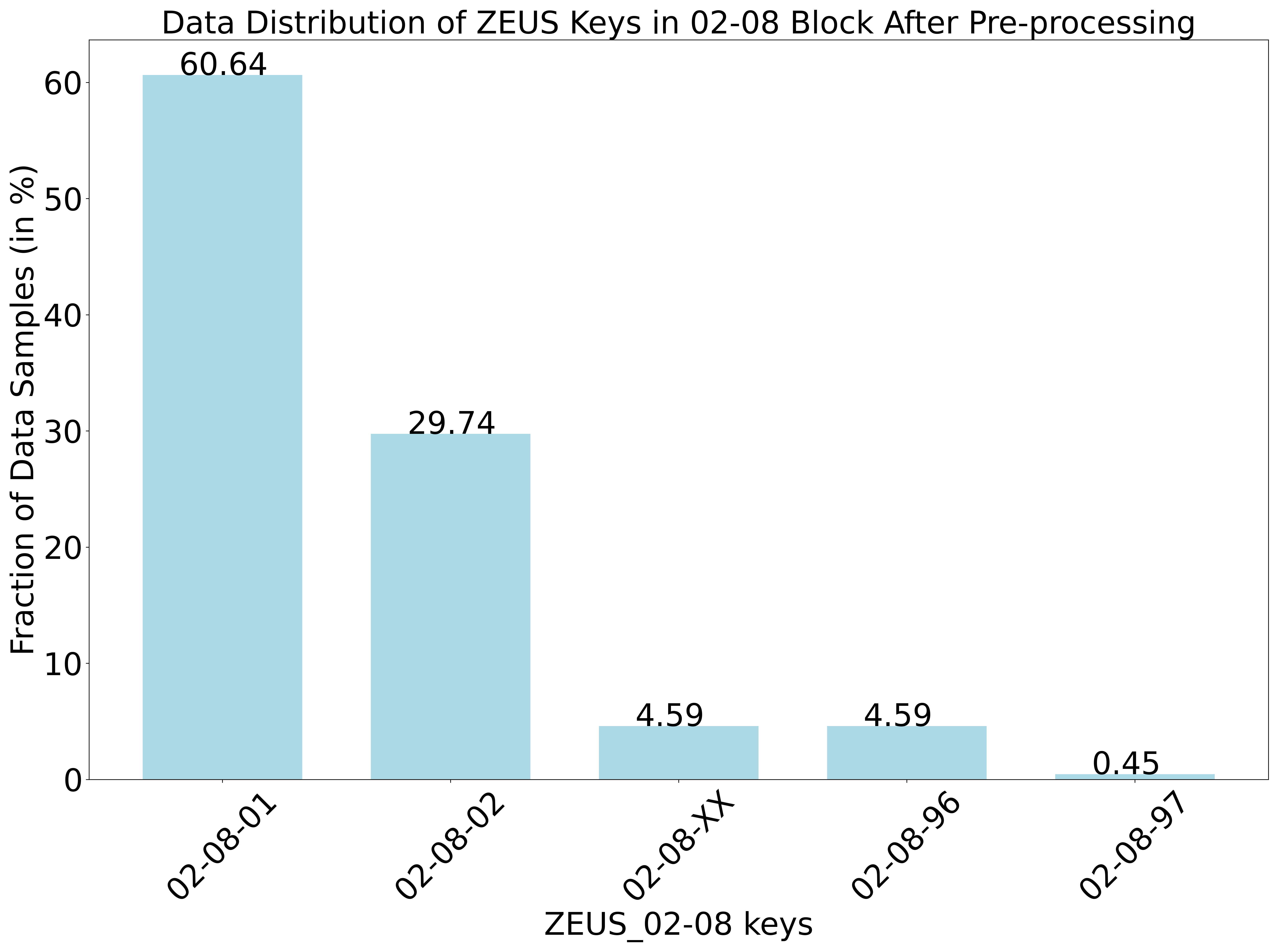}
    \caption{Data distribution of \ac{zeus} classes after \ac{mwo} preprocessing. \added{For classification, \ac{zeus} level 3 is considered. Classes are distributed highly unevenly.}}
    \label{fig:preprocessing}
\end{figure}

The data set consists of MWOs from forty \acp{wt} belonging to a wind farm. The samples were collected between January 2016 and December 2020, spanning a time window of 4 years. A total of 3896 MWOs were collected from all the forty \acp{wt}.
The data distribution of the MWOs into \ac{zeus} classes is depicted in Figure \ref{fig:preprocessing}. More than 90\% of the data samples belong to the \ac{zeus} classes 02-08-01 \added{(corrective maintenance)} and 02-08-02 \added{(preventive maintenance)}. The remaining \acp{mwo} belong to the other level three \ac{zeus} classes. This constituted the raw data after preprocessing it for training and tagging.

\begin{comment}

\begin{figure}
\centering
\begin{subfigure}{.5\textwidth}
  \centering
  \includegraphics[width=\linewidth]{Definitions/data_distribution_level_three_before_preprocessing.png}
  \caption{Before Preprocessing}
  \label{fig:beforepreprocessing}
\end{subfigure}%
\begin{subfigure}{.5\textwidth}
  \centering
  \includegraphics[width=\linewidth]{Definitions/data_distribution_level_three_after_preprocessing.png}
  \caption{After Preprocessing}
  \label{fig:afterpreprocessing}
\end{subfigure}
\caption{Data Distribution in ZEUS Classes}
\label{fig:preprocessing}
\end{figure}
\end{comment}

The preprocessing of texts in the \acp{mwo} consisted of the following steps \cite{hotho2005brief,denny2018text}.
\begin{itemize}
    \item Lower case conversion.
    \item Removal of white spaces.
    \item Removal of punctuation.
    \item Removal of numbers.
    \item Tokenization.
    \item Removal of unimportant words.
    \item Removal of stop words.
    \item Dropping of empty rows.
\end{itemize}

Figure \ref{fig:preprocessing} shows the data distribution after preprocessing. \deleted{Once we observe the percentage distribution value for the class 02-08-02, we noticed that} There is a slight decrease in the amount of data points belonging to the class 02-08-02, since \added{empty reports} \deleted{they} were dropped as part of the preprocessing step \footnote{Note that the values are rounded to the second decimal. Therefore the sum of the bars is greater than 100\%.}. This\added{, in turn,} resulted in a slight \added{percentage} increase \deleted{in ratio} for the data points belonging to the classes 02-08-01, 02-08-XX and 02-08-96 respectively.

\section{Results}
\label{results}
This section presents the failure rate \ac{kpi} \added{computation} results of each approach and compares them against each other.

\subsection{Expert Labeling}
For the ground truth labels resulting \deleted{in} \added{from} the first approach, \ac{mtbf} is calculated for all five level 3 classes of \ac{zeus} block 02-08 by considering the difference between the start dates for subsequent \acp{mwo}. As the mean time between subsequent events are different for different maintenance types (e.g., corrective maintenance or preventive maintenance), the resulting \ac{mtbf} values are different for each \ac{zeus} class. Next, the maintenance event rate values are calculated for each class by the inverse of the \ac{mtbf} values. As only the corrective maintenance tasks are associated with the failure, the corrective maintenance class (02-08-01) is considered for the calculation of \added{the} failure rate \ac{kpi}. The failure rate  for the 02-08-01 class is 8.85~1/a. It took 231 hours in total for the experts to tag all the \ac{mwo} samples. \deleted{and thus the total tagging effort is 231 hours for this method.} Next to \ac{zeus}, which identifies the state of the \ac{wt}, also the component described in the \ac{mwo} has been labeled within this time period according to a domain standard. Only the total tagging time is assessed by the authors, within this time the tagging with \ac{zeus} has the biggest share.

\subsection{Automated classification according to technical guidelines}

\begin{table}
\centering
\caption{Comparison of ZEUS classifier performance. \added{We depict macro average and weighted averages for precision, recall and $F_1$-Score. Results are given for logistic regression (LR) and naive Bayes (NB) classifiers. Training data was oversampled using either \ac{smote} \cite{smote} or \ac{ro} \cite{he2009learning}}}\label{table:classifiercomparison}
\begin{tabular}{l l l l l}
\toprule
\textbf{Classifier} & & \textbf{Precision} & \textbf{Recall} & \textbf{$F_1$-Score}\\
\midrule
    \multirow{2}{*}{\textbf{NB+RO}} & \textbf{Macro Avg} & 0.65 & 0.75 &  0.67 \\
        & \textbf{Weighted Avg} & 0.84 & 0.77 & 0.80 \\
        \midrule
    \multirow{2}{*}{\textbf{NB+SMOTE}} & \textbf{Macro Avg} & 0.68 & 0.74 & 0.70 \\
    & \textbf{Weighted Avg} & 0.84 & 0.81 & 0.82\\
    \midrule
    \multirow{2}{*}{\textbf{LR+RO}} & \textbf{Macro Avg} & \textbf{0.76} & \textbf{0.74} & \textbf{0.75} \\
    & \textbf{Weighted Avg} & \textbf{0.85} & \textbf{0.85} & \textbf{0.85} \\
    \midrule
    \multirow{2}{*}{\textbf{LR+SMOTE}} & \textbf{Macro Avg} & 0.69 & 0.74 & 0.71 \\ 
    & \textbf{Weighted Avg}  & 0.85 & 0.84 &0.84 \\ 
    % Added LLM results (distilRoberta_epochs_20-LR-1e-06)
    \bottomrule
%    \multirow{2}{*}{\added{\textbf{DistilRoBERTa}}} & \textbf{Macro Avg} & 0.67 & 0.60 & 0.63 \\ 
%    & \textbf{Weighted Avg}  & 0.78 & 0.80 & 0.79 \\ 
%    \bottomrule
\end{tabular}
\end{table}

\begin{table}
    \centering
    \caption{Detailed classification results of \added{the best performing model} LR+RO. \added{Next to macro and weighted averages, we also give scores for each \ac{zeus} class. \ac{zeus} class 02-08-97 was removed since there were too few samples.}}
    \begin{tabular}{l l l l l}
    \toprule
        \textbf{ZEUS\_02-08} &     \textbf{Precision} & \textbf{Recall} & \textbf{$F_1$-Score} & \textbf{Support} \\ \midrule
        \textbf{02-08-01} & 0.89 & 0.90 & 0.90 & 0.61 \\
        \textbf{02-08-02} & 0.84 & 0.85 & 0.84 & 0.30 \\
        \textbf{02-08-96} & 0.40 & 0.47 & 0.43 & 0.04 \\
        %\textbf{02-08-97} & n/A & 0.00 & n/A & 0.00 \\ \hline
        \textbf{02-08-XX} & 0.92 & 0.74 & 0.82 & 0.05 \\ 
        %\textbf{Accuracy} & ~ & ~ & 0.85 & 1.00 \\ \hline
        %\textbf{Macro Avg} & 0.61 & 0.59 & 0.60 & 1.00 \\
        \midrule
        \textbf{Macro Avg} & 0.76 & 0.74 & 0.75 & 1.00 \\
        \textbf{Weighted Avg} & 0.85 & 0.85 & 0.85 & 1.00 \\ \bottomrule
    \end{tabular}
\label{fig:Classification Results of LR+RO}
\end{table}

\begin{figure}
\begin{center}
\includegraphics[width=0.8\textwidth]{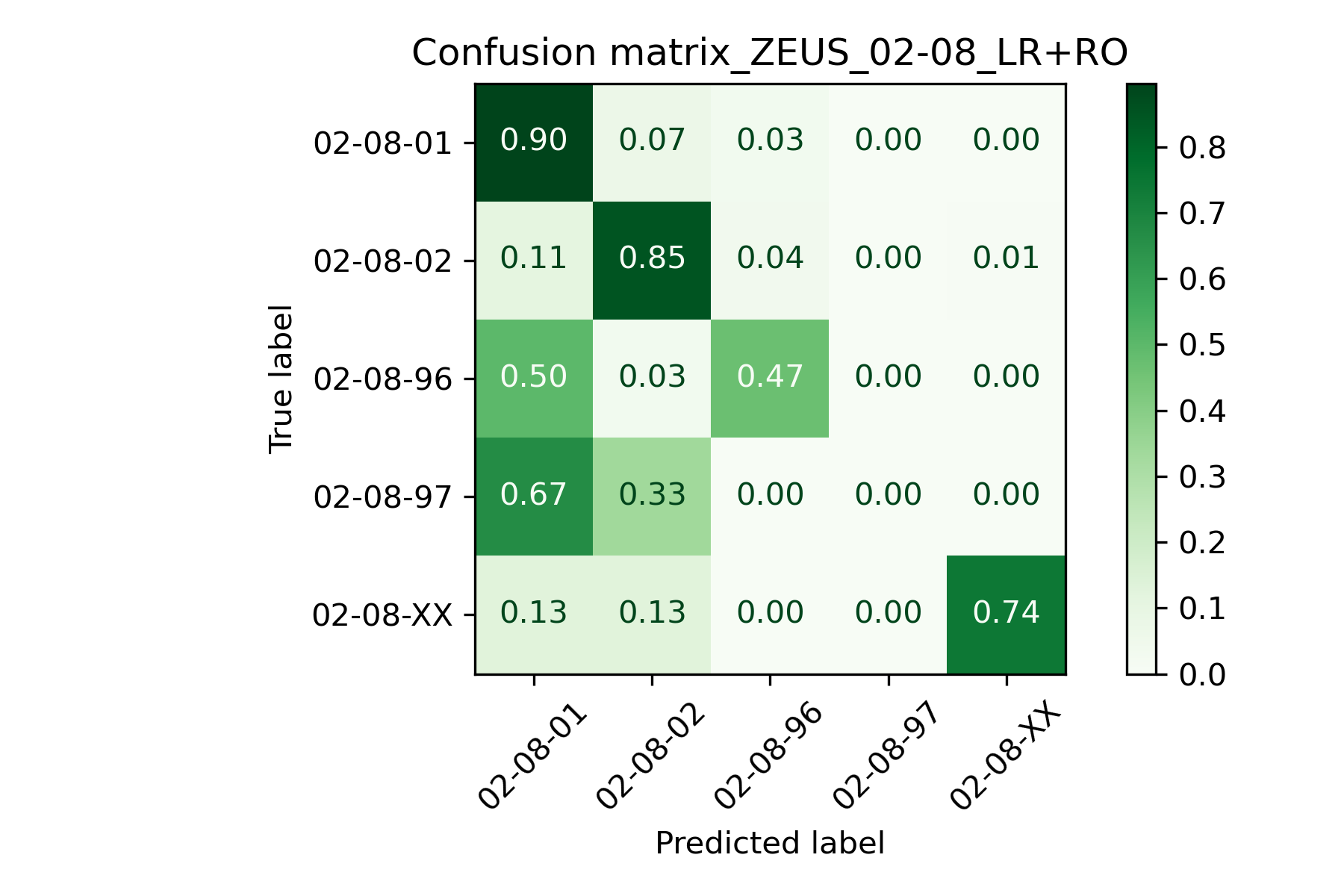}
\caption{Confusion matrix \added{for the best performing model} \ac{lr}+\ac{ro}. \added{Classes with few samples pose a principal challenge in \ac{mwo} classification.}}\label{fig:Confusion Matrix - LR+RO}
\end{center}
\end{figure}

The failure rate \ac{kpi} for the second approach is calculated as indicated \deleted{by} \added{in} step 5 of Figure \ref{fig:Methodology}. Multiple \ac{ml} models were developed as part of this step.  The macro average precision, recall and $F_1$ score (i.e., average over all classes, \cite{takahashi_confidence_2022}) and weighted averages (i.e., averages weighted by the respective relative class frequencies) for all classification and oversampling method combinations are given in Table \ref{table:classifiercomparison}. The best performance values are highlighted in bold. For the calculation of these values, we removed the \ac{zeus} class 02-08-97, since in three out of the four cases considered our classifiers did not predict any test sample to belong to this class. Hence, although oversampling methods were used in the training, calculation of precision and $F_1$ scores for this class as well as the calculation of averages over all classes was not possible.

\added{We may note that we also conducted a first experiment using a transformer-architecture-based \cite{vaswani2017attention} large language model (LLM), namely DistilROBERTa \cite{roberta}, since LLMs constitute the current state-of-the-art in text classification. However, as expected, due to the small amount of training data, this model was merely able to achieve macro average $F_1$-Scores of 0.63 and a weighted average of 0.79, i.e., lower scores than both of the more traditional text classification methods. We suspect that, in the long run, better performance in \ac{mwo} classification can be achieved through LLMs. However, they require additional optimization for the application to technical language and more elaborate methods for handling minority classes with few samples, such as domain adaptation, transfer learning \cite{transferlearning} and active learning \cite{activelearning}. }

For \ac{kpi} calculation, the best performing model was considered, \deleted{The best performing model} \added{which} is \ac{lr} \deleted{when} \added{with} \ac{ro} \deleted{is} used for oversampling the minority class data. \added{Somewhat surprisingly, random oversampling mostly leads to better scores than the more elaborate SMOTE for both classification models.} The detailed classification results of the \ac{lr}+\ac{ro} model including performances for each class are \deleted{as observed} \added{given} in Table \ref{fig:Classification Results of LR+RO}. \deleted{and Figure \ref{fig:Confusion Matrix - LR+RO.}} Figure \ref{fig:Confusion Matrix - LR+RO}\added{, which} depicts the confusion matrix for the \ac{lr}+\ac{ro} classifier. \deleted{We can observe from the figure} \added{It shows} that nearly 90\% of the data samples belonging to the ZEUS class 02-08-01 and 85\% of the data samples belonging to the ZEUS class 02-08-02 have been classified correctly. In contrast to this, \deleted{the} scores are lower for the remaining classes, possibly since they are minority classes and constitute only 10\% of the total sample, \deleted{i.e.,}\added{which means that} less training data was available. We may note that \deleted{overall} \added{in general,} except \deleted{for}\added{from} ZEUS class 02-08-97 the scores are considerably higher than the expected values that would result from a random classification (i.e., \deleted{which predicts} each class \added{being predicted} either with uniform random probability or with probabilities proportional to the class distribution). %We may note that results for the \ac{zeus} class 02-08-97 are not present in Table \ref{tab:zeus block 02-08} since this class was never predicted by the \ac{lr}+\ac{ro} classifier and hence calculation of precision and f1 score is not possible. %We may note that only three samples exist in the data with \ac{zeus} class 02-08-97 as their ground truth label. Furthermore, the \ac{lr}+\ac{ro} method did not predict any sample from the test set to belong to this class. For this reason, the support of this class rounded to two decimal places is 0.00 and the precision and F1 score are undefined.

The \ac{lr}+\ac{ro} model outperformed other \ac{ml} models with an overall accuracy of 85\% and thus, the tagged data resulting from this model are considered for \ac{kpi} calculation. \deleted{As this approach followed} \added{Following} the \ac{zeus} guideline, the \ac{kpi} calculation steps are the same as in the first approach \added{(expert labeling)}. The event rate values were calculated for all \deleted{the five} \ac{zeus} classes. \added{As mentioned before, } the maintenance event rate for the 02-08-01 class is \added{considered as} the failure rate and the \added{average} value obtained \added{over all \acp{wt}} is 8.21~1/a. As \added{the text classification}\deleted{this} method uses manually labeled \acp{mwo} from the domain experts \added{for training and testing}, the tagging time is still 231 hours. However, \deleted{this}\added{labeling} only needs to be \deleted{applied}\added{conducted} once, thus saving many hours across the remaining life cycle of \deleted{these units} \added{the \acp{wt}}.

\subsection{Human-in-the-Loop Tagging}
The failure rate calculation for the third method \deleted{followed}\added{is based on} a different approach since Nestor tagging does not follow the \ac{zeus} guideline. As four different rules are considered, the \added{\ac{mtbf}} value\deleted{s} \deleted{\ac{mtbf} of obtained is different for each rule} \added{may also be considered for each rule by itself}. This results in four separate maintenance event rate values. Out of all the four rules, the fourth rule considers all the \acp{mwo} in the P entity and \deleted{also successively filters out the} \added{excludes} descriptions \deleted{where}\added{in which} the P entity tag is present but \added{indicates the absence of a failure.} \deleted{the description indicates no failure}. Thus, \added{we assume that} Rule 4 \deleted{is believed to be}\added{is} most robust. The failure rate \deleted{value for this method}\added{computed from this rule} is 6.89~1/a.

\subsection{Comparison for Reliability KPI calculation}
\begin{table}%[!ht]
    \centering
    \caption{Comparison of KPI results \added{of the three approaches. Strathclyde results are added for additional comparison as reported in \citet{carroll2016failure}.}}
% \usepackage{multirow}
\begin{tabular}{l l l l | l}
\toprule
\textbf{}                                                                  & \begin{tabular}[c]{@{}l@{}}\textbf{Expert Labeling}\end{tabular} & \textbf{LR+RO}                                            & \textbf{Nestor} & \textbf{Strathclyde} \cite{carroll2016failure} \\
\midrule
\textbf{\begin{tabular}[c]{@{}l@{}}Failure Rate\\ {[}1/a{]}\end{tabular}}  & 8.85                     & 8.21                                                      & 6.89            & 8.27          \\
\textbf{\begin{tabular}[c]{@{}l@{}}Tagging Time\\{[}h{]} \end{tabular}} & 231                      & \begin{tabular}[c]{@{}l@{}}231\\ (initially)\end{tabular} & 28              & N/A \\ \bottomrule
\end{tabular}
\label{fig:Comparision of KPI Results}
\end{table}

As part of the final step 10 in Figure \ref{fig:Methodology}, the \ac{kpi} results of the second and third approach are compared against the results of the first approach. Table \ref{fig:Comparision of KPI Results} shows the results. The failure rate value for the first approach is 8.85~1/a and this is considered to be the ground truth value as it is derived from the data that is tagged by the domain experts. For the second technique, the \ac{kpi} results of the \ac{lr}+\ac{ro} model is considered as it is the best performing \ac{ml} model. The failure rate value for the \ac{lr}+\ac{ro} model is 8.21~1/a. \deleted{If we compare this value with the failure rate calculated from the first approach, the values are close to each other and thus}\added{Since this value is close to the ground truth} we \deleted{can} conclude that the second method yields reliable results. \deleted{If we consider}\added{Considering} the tagging time, \ac{lr}+\ac{ro} model uses the \deleted{tagged data from the}\added{data labeled by the} experts in the first approach \added{for training}\deleted{to train} the \deleted{\ac{ml}}\added{classification} models. Hence, the tagging time is the same for both, i.e., 231 hours. For \deleted{the} Nestor tagging, the fourth rule was considered for the calculation of failure rate as it is the most robust of all \added{introduced} rules\deleted{ considered}. The failure rate value obtained from \deleted{the third}\added{this} approach is 6.89~1/a. When comparing the \ac{kpi} results of the first and the third approach, we observe that the results are \deleted{not close as}\added{less accurate than} with the second approach. One possible reason for this is the different tagging and \ac{kpi} calculation approach resulting from the Nestor tool. However, \deleted{if we} consider\added{ing} the \deleted{time taken for} tagging \added{time} \deleted{which is just} \added{of only} 28 hours for Nestor, the \ac{kpi} results are still \deleted{decent}\added{valuable to obtain a rough estimate with considerably less effort}. The resulting 88\% drop in manual labor hours provided a result within 22.5\% of our ground truth value, despite the fact that Nestor tool used does not directly align with the \ac{zeus} criteria. From these results we could expect that developing a  human-in-the-loop tool like Nestor which is aligned with the \ac{zeus} criteria could save thousands of man-hours while maintaining high levels of accuracy. \added{The accuracy might benefit from further optimization of tagging procedures and extraction rules in future work.}

\section{Discussion}
\label{discussion}
From the results of section \ref{results}, it is evident that in addition to expert labeling using the \ac{zeus} approach, automatic classification through \ac{ml} and Nestor tagging can also be applied to extract reliability \acp{kpi} from the \ac{wt} \acp{mwo}. If we consider the quality of the \ac{kpi} results in Table \ref{fig:Comparision of KPI Results}, it is evident that the second approach results are very close to the original results and thus automatic classification using the \ac{lr}+\ac{ro} model is the most \deleted{suitable}\added{precise automated} method \deleted{to}\added{for} extract\added{ing} reliability \acp{kpi}. Nestor tagging on the other hand is still reliable \added{enough for a rougher \ac{kpi} estimate} considering that it helps to save 88\% of manual tagging effort. Table \ref{fig:Comparision of KPI Results} also contains the failure rate results of an additional initiative namely Strathclyde, which was mentioned in section \ref{sota}. \added{Interestingly, the calculated failure rate from this initiative is very similar to that of our text classification method, while the expert labeling approach, which served as our ground truth, leads to a slightly higher value. The reason for this may lie either in a similar accuracy of the method, but also in the difference of the data set.} \deleted{This is an external approach and the results are from different research work on a completely different data set. The results from the external work are included only for comparison purpose and are completely independent from our methodology.}

\added{Based on our overall results, we are able to derive the following recommendations for KPI extraction from \ac{mwo}s: To achieve the most accurate results, automated text classification should be used. In order to spare a considerable amount of working hours while achieving a coarser estimate of KPIs, AI-assisted tagging is the appropriate approach. Random Oversampling can be used to improve \ac{mwo} classification performance for less frequent classes.}

\section{Conclusion}
\label{conclusion}
\deleted{Planning and executing the most efficient maintenance operations for \ac{wt} pose many challenges because one of the best sources of information about this, \acp{mwo}, are often unstructured and non-standardized in their collection making them difficult to analyze. The purpose of this work is to emphasize the importance of enabling, accelerating, and following the use of standards and guidelines in the \ac{wt} industry. The three approaches described in the methodology enable the optimization of \ac{wt} maintenance activities based on \acp{kpi}.}

\added{In the present work,} we exploit\added{ed} the benefits of supervised \acp{ml} techniques to automatically classify \ac{mwo} text descriptions into their respective \ac{zeus} classes. We also investigated a human-in-the-loop Nestor tagging approach where single words were individually tagged and analyzed. \added{Based on the ground truth derived from a manual expert labeling approach, we have compared the methods in terms of KPI calculation accuracy as well as tagging time. Our overall results indicate complementary strengths and weaknesses of the methods: Text classification leads to accurate KPI estimation, while tagging time for the creation of training data is very high. AI-assisted tagging through Nestor, while being slightly less accurate, leads to greatly reduced tagging times.}

\added{The used approaches also come with technical limitations. The used text classification methods do consider the frequency, but not the sequential order of tokens within each \ac{mwo}. Furthermore, they are not able to distinguish between homonyms. A more elaborate method such as a large language model would improve upon this. Our first experiments in this field using DistilRoBERTa suffer from too few training data. However, we assume that LLMs and deep learning techniques bear great potential when further optimization for technical language and dealing with few training data is carried out, e.g., through transfer learning \cite{transferlearning} or active learning \cite{activelearning}. We reserve further investigations within this area for future work.}
\deleted{This research can be continued further by applying more elaborate \acp{ml} concepts such as those based on deep learning methods.}\added{As a further direction,} future research will explore the potential of using tools such as Nestor, and more elaborate \ac{ml} methods to tag multiple words at once, improve tagging accuracy, and aid in diagnostic and predictive corrective action decision making.

\authorcontributions{Marc-Alexander Lutz: Methodology,  validation, writing; Bastian Schäfermeier: Methodology,  validation, writing; Rachael Sexton: Software, writing---review and editing; Michael Sharp: Software, writing---review and editing; Dima Alden: Software, writing---review and editing; Stefan Faulstich: writing---review and editing;  Jagan~Mohini Aluri: Methodology,  validation, writing ;\\
All authors have read and agreed to the published version of the manuscript
}

\funding{This research is to a large part carried out within the research project "Digitalization of
Maintenance Information" (DigMa) funded by the German Federal Ministry of Economic Affairs
and Climate Action (BMWK), grant number 03EE2016A.}

\dataavailability{Data is not available due to privacy restrictions.}

\acknowledgments{The authors thank the project
partners for providing field data and for sharing their requirements and experience.}

\conflictsofinterest{The authors declare no conflict of interest. The funders had no role in the design of the study; in the collection, analyses, or interpretation of data; in the writing of the manuscript; or in the decision to publish the results.} 
\newpage
%%%%%%%%%%%%%%%%%%%%%%%%%%%%%%%%%%%%%%%%%%
\begin{adjustwidth}{-\extralength}{0cm}
%\printendnotes[custom] % Un-comment
\reftitle{References}
%\bibliography{literature}

\end{adjustwidth}

\end{document}